\documentclass[sigconf]{acmart}

\usepackage{booktabs} 

\setcopyright{rightsretained}

\acmDOI{10.475/123_4}

\acmISBN{123-4567-24-567/08/06}

\acmConference[WOODSTOCK'97]{ACM Woodstock conference}{July 1997}{El
  Paso, Texas USA}
\acmYear{1997}
\copyrightyear{2016}

\acmArticle{4}
\acmPrice{15.00}

\editor{Jennifer B. Sartor}
\editor{Theo D'Hondt}
\editor{Wolfgang De Meuter}

\copyrightyear{2018}
\acmYear{2018}
\setcopyright{acmlicensed}
\acmConference[AIES '18]{2018 AAAI/ACM Conference on AI, Ethics,
and Society}{February 2--3, 2018}{New Orleans, LA, USA}
\acmBooktitle{2018 AAAI/ACM Conference on AI, Ethics, and Society
(AIES '18), February 2--3, 2018, New Orleans, LA, USA}
\acmPrice{15.00}
\acmDOI{10.1145/3278721.3278744}
\acmISBN{978-1-4503-6012-8/18/02}

\begin{document}
\title{Towards Composable Bias Rating of AI Services}

\author{Biplav Srivastava}
\affiliation{%
  \institution{IBM Research}
  \streetaddress{IBM T. J. Watson Research Center}
  \city{Yorktown Heights}
  \state{NY}
  \postcode{10584}
}
\email{biplavs@us.ibm.com}

\author{Francesca Rossi}
\affiliation{%
  \institution{IBM Research}
  \streetaddress{IBM T. J. Watson Research Center}
  \city{Yorktown Heights}
  \state{NY}
  \postcode{10584}
}
\email{Francesca.Rossi2@ibm.com}

%
%
%
%

\renewcommand{\shortauthors}{B. Srivastava and F. Rossi}

\begin{abstract}

A new wave of decision-support systems are being built today using AI services that draw insights from data (like text and video) and incorporate them in human-in-the-loop assistance. However, just as we expect humans to be ethical, the same expectation needs to be met by automated systems that increasingly get delegated to act on their behalf. A very important aspect of an ethical behavior is to avoid (intended, perceived, or accidental) bias. Bias occurs when the data distribution is not representative enough of the natural phenomenon one wants to model and reason about. The possibly biased behavior of a service is hard to detect and handle if the AI service is merely being used and not developed from scratch, since the training data set is not available. In this situation, we envisage a 3rd party rating agency that is independent of the API producer or consumer and has its own set of biased and unbiased data, with customizable distributions.
We propose a 2-step rating approach that generates bias ratings signifying whether the AI service is unbiased compensating, data-sensitive biased, or biased. The approach also works on composite services. We implement it in the context of text translation and report interesting results.  

\end{abstract}

%
%



\maketitle


\section{Introduction}

The popular approach for building software applications today is by reusing any existing capability from others exposed as Application Programming Interfaces (APIs), and developing new code for the rest, as well as glue code to connect them \cite{api-hype}. Service catalogs facilitate API discovery by enabling search by an API's functional (e.g., description) and non-functional capabilities (e.g., cost, availability).  Most API catalogs, whether public, like ProgrammableWeb \cite{pw}, or private by cloud vendors, list services based on metadata and cost. 
As  adoption of such AI services increases that draw insights from data and get incorporated into the human-in-the-loop decision-making, the expectation of ethical decisions from humans gets extended to automated systems that increasingly get delegated to act on their behalf, or that recommend decisions to humans. 

There are many aspects of an ethical behavior that we expect from a decision making entity. Prominent among them are alignment to common norms, transparency, fairness, diversity, and interpretability.
In particular, fairness refers to the behavior that treats all elements of a certain class in the same way. A more precise term for fairness is bias. In an ethical system, it is important to avoid behaving in a way that presents intended, perceived or accidental bias. 


\begin{figure*}[ht]
    \centering
   \includegraphics[width=0.75\textwidth]{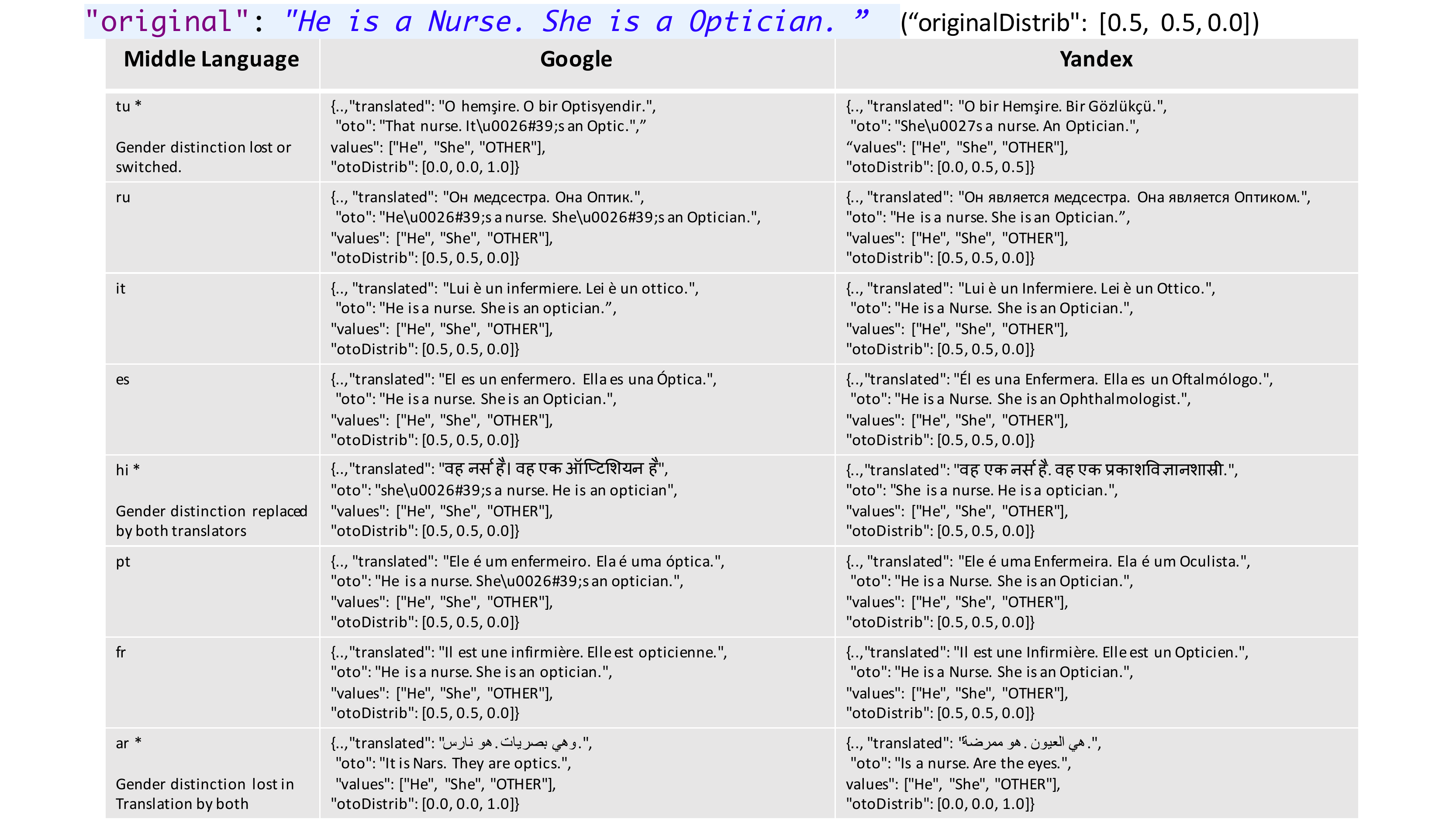}
    \caption{Example of gender distribution distortion when translating from English $\rightarrow M_i \rightarrow $ English with two public translation APIs. Note: $\rightarrow$ refers to one translation, $M_i$ is a middle language, and  {\em oto} refers to output English text. Accessed: Nov 14, 2017.}
    \label{fig:example}
\end{figure*}


More precisely, bias occurs with respect to an attribute (such as gender or race) when the data distribution is not representative enough of the natural phenomenon (that is, the distribution of the attribute's values) that one wants to model and reason about. For example, if we search for images of engineers in ImageNet, we will get very few womens, in percentage which is much lower than the actual percentage of women engineers in real life. If such dataset would be used to train a system that is intended to make decisions (or help humans make decisions) about engineers, the system would possibly not treat women engineers in a fair way. Bias has been shown in many existing AI systems that are currently used, for example in the algorithm used by the US judicial system to predict which criminals have a high probability of reoffending, which has been shown to be biased against African Americans \cite{propublica}.

If the dataset used for training the system is available, it is easy to check if it is biased, and there are technical solutions that allow to partially remove the bias. However, if the AI service is merely being used by a consumer and not developed from scratch, so the training data set is not available, the possibly biased behavior of the service is hard to detect and handle.

In this paper we consider this scenario and  show how to detect bias through a two-step test approach, and to rate the AI service according to the kind of bias that has been recognized. We consider bias as any abnormal distribution of values of an attribute from one or more baseline distributions that are considered unbiased (or normal). For example, the attribute {\em Gender} may have values {\em He}, {\em She} and {\em Other}, and attribute {\em Place of Worship} may have attribute values {\em Church, Mosque, Temple, Synagogue, Other}. We will focus on gender for the rest of the paper but the discussion applies to any attribute of interest.

\subsection{Illustration and Running Example}

Let us consider a simple hypothetical AI system called {\em UniversalSocialRepeater} (USR) that takes an English input text from a person 
and posts it online to a forum like Twitter and Facebook in multiple world languages of their choice. The output (translated post) is read by other people who may be offended if the person ($p_i$) is perceived to act in a biased way. If the translated text expresses bias, the user may have written a biased English text and that bias has been propagated to the translated version of the text, or the application (USR) has introduced bias. USR itself can be offered as an API service to be used by other developers. 

Suppose a startup wants to build USR using off-the-shelf translation AI services, denoted $A^t : l_i \rightarrow l_j$, 
where  $l_i$ and $l_j$ are natural languages and $\rightarrow$ is the supported translation direction. Examples of $A^t$ APIs are Google Translate \cite{google-translate} ($A^t_g$) and Yandex  Translate \cite{yandex-translate}($A^t_y$). 

Today, most API catalogs list services based on metadata and cost
but do not define the bias they may introduce. 
The bias rating we are proposing will be very important for the startup to demonstrate that it is not introducing bias on its own, for example by building a test service, say {\em USRTest} for this example, which will translate the user's 
English text to a specified language (called middle language $M_i$), and then back to English, so that the user can verify any attribute's data distributions and themselves verify the system behavior. That is, USRTest = ($A^t_j$ : $E \rightarrow M_i$) $\star$ ($A^t_j$ : $M_i \rightarrow E$), where $A^t_j$  is a translator, E is English and $\star$ denotes composition.  

We illustrate the gender distortion for such a USRTest scenario in Figure~\ref{fig:example}. The input English text was - {\em He is a Nurse. She is a Optician}. It was run for 8 middle languages and the two translators. We notice that not only gender distinction is lost when $M_i$ is Arabic or Turkish, but also that it is even switched when the middle language is Hindi. 
It is important to notice that not only the gender bias will occur when one uses any of the two translators, but it will also be progatated by downstream applications built consuming $A^t_g$ and $A^t_y$.

\subsection{Rating of AI Services}

In situations like the one described in the above example, we envisage a 3rd party rating agency that is independent of the API producer or consumer and that has its own set of biased and unbiased data, with customizable distributions. 

Given an API, an attribute, and a declaration of what it means for a dataset to be biased (or unbiased) regarding that attribute, we propose a 2-step rating approach that generates a 3-level bias rating, signifying whether the AI service is unbiased compensating (UCS, which means it does not introduce bias and can even compensate for a possibly biased data set), data-sensitive biased (DSBS, meaning that the API follows the bias properties of the input dataset), or biased (BS, meaning that the API may introduce bias even when the dataset is unbiased).
 
We perform an extensive experimental analysis on translation services, checking possible gender bias in going from English to English via a second language. We consider two translation APIs (Google and Yandex) and eight middle languages,
comparing the two services in terms of their capability to avoid bias.

Our approach also works on composite services. 
This is crucial, since most services can be obtained by composing simpler steps. An example is a service that takes an image and generates a sentiment, which can be obtained by composing sequentially a service that takes an image and generates a caption, and another service that takes the caption and generates a sentiment.

In summary, the main contributions described in this paper are as follows:
\begin{itemize}
\item The definition of a 2-step testing that takes in input an API, a protected variable, and bias/unbiased distributions, and returns a bias rating.
\item The proposal and discussion of a 3-level rating of AI services based on distribution distortion. 
\item The implementation of a 3rd-party rating of AI services.
\item An extensive experimental analysis that uses commercial text  translation services and derive new insights on the bias-related behavior of the services.
\item The definition and analysis of the properties of the sequential compsition of the proposed bias rating approach.

\end{itemize}


\subsection{Structure of the Paper}

In the rest of the paper, we begin with relevant background on ethics and bias in AI services, and then present our 
procedure for 
rating AI services along a 3-level bias scale. We then present an implementation of our approach with focus towards  text translation for the USRTest task. We conduct experiments with 2 commercial translators and 8 middle languages, and conclude with a discussion of key insights, limitations and future work. 





\section{Computational Ethics and AI Bias}

Humans are usually social agents who live in a community, and ethics and morality are ways to guide our behavior so that both social and individual wellbeing is coherently achieved and maintained. Therefore we  usually constrain our decisions according to moral or ethical values that are suitable for the scenario in which we live. In the same way, AI systems that have an impact on real life environments or on humans, or that recommend decisions to be made by humans, should be designed and developed in a way that they follow suitable ethical principles as well.
This is why ethical decision making has been widely studied in AI, to understand how to teach an AI system to act within ethical or legal guidelines (see for example \cite{moral-machines}).
%

While ethical principles are not universal and can vary according to scenarios, tasks, domains, and cultures,  one property that is usually included in the realm of ethical behavior is fairness, that is, 
the impartial and just treatment or behavior without favoritism or discrimination. The absence of fairness is usually referred to as bias.
Since AI systems are increasingly acting in the real world, they should be fair as well. Thus it is important to understand how to recognize possible bias in AI services, or even to eliminate it. 

Many have already considered this task. For example, in \cite{video-bias}, the authors look at gender issues during structured prediction for vision recognition. Also, in \cite{kush}, the authors propose a method to eliminate bias in a dataset when focussing on a specific protected attribute (such as gender or race), while trying to maintain the same distribution for all other predictive attributes.



Indeed, bias in AI services can be exposed in many ways.
The input data that is used to train the AI system may 
present bias on some attribute, such as gender or race.
Also, the data may be fair but the learning algorithm could introduce bias.
When the training data is available, one can examine it and try to possibly recognize and remove the bias, making it fairer. This is the approach followed in \cite{kush}. 

However, when the service is just used but the training data is not available, this examination cannot be perfomred, so the only approach is to test the service against bias.
This is the approach we take in this paper.





\section{Bias Testing}

\subsection{Bias Rating for AI Services}

\begin{figure}[ht]
    \centering
   \includegraphics[width=0.4\textwidth]{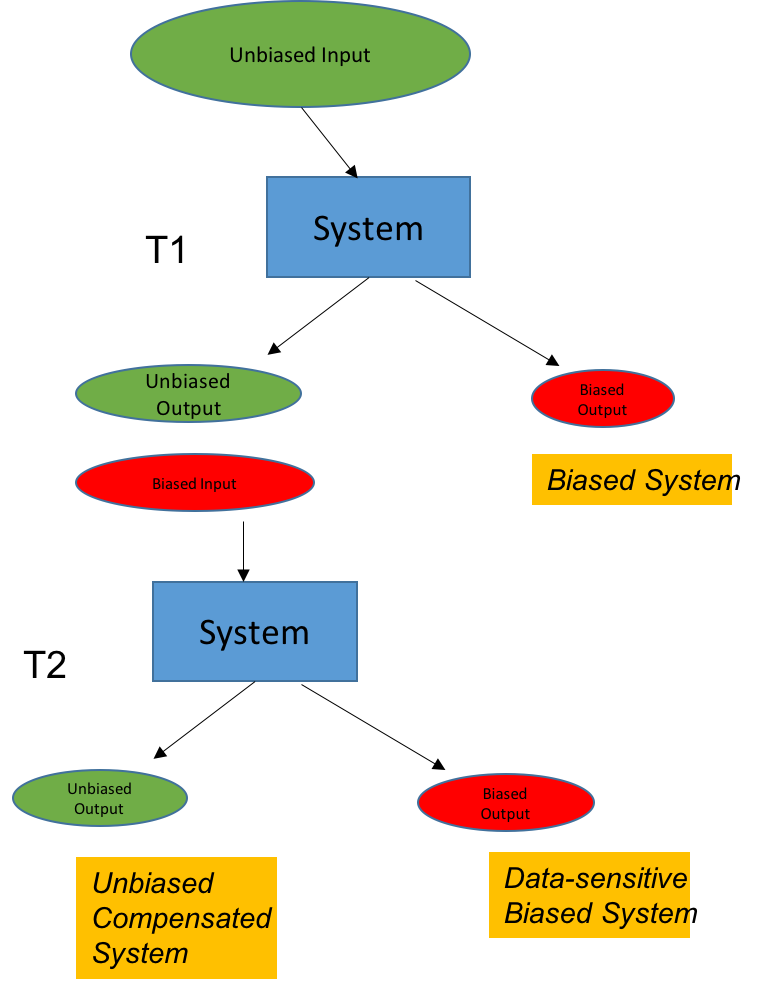}
    \caption{The 2-step approach to rate an AI service, shown as a decision tree.}
    \label{fig:dec-tree}
\end{figure}

We propose a procedure for rating an AI service against bias, as shown in Figure~\ref{fig:dec-tree}. In the first stage (T1), the AI system under consideration is subjected to unbiased input and its output is analyzed. If the output is biased, the system is rated Biased (BS) under that test.  This means that the system introduced bias even when the input is unbiased. This is the worst rating that could come out of the procedure.


If on the contrary the output is unbiased, the system is now subjected to biased input (T2) and its output analyzed. If the output is biased, the system is rated Data-Sensitive Biased System (DSBS). This means that 
the system does not introduce bias but it follows whatever bias is present in the input. This is a negative rating but at least we know that, if we could remove bias from the input data, the system would not introduce it.

If instead the output is unbiased, given biased input in the testing stage T2, then we output the rating Unbiased Compensated System (UCS). 
This is the best rating and it means that the system not only does not introduce bias, but it does not even follow the bias of the input data,
and is instead able to compensate for possible bias in the input data.

Note that the test could also have been done in the reverse manner by starting with biased input. But in that case, both cases of biased or unbiased outputs would need a second test each, with unbiased input data, to give a rating. Due to the extra cumulative testing needed (3 v/s 2), we do not follow that path.

\subsection{Architecture of our Rating System}

\begin{figure}[ht]
    \centering
   \includegraphics[width=0.5\textwidth]{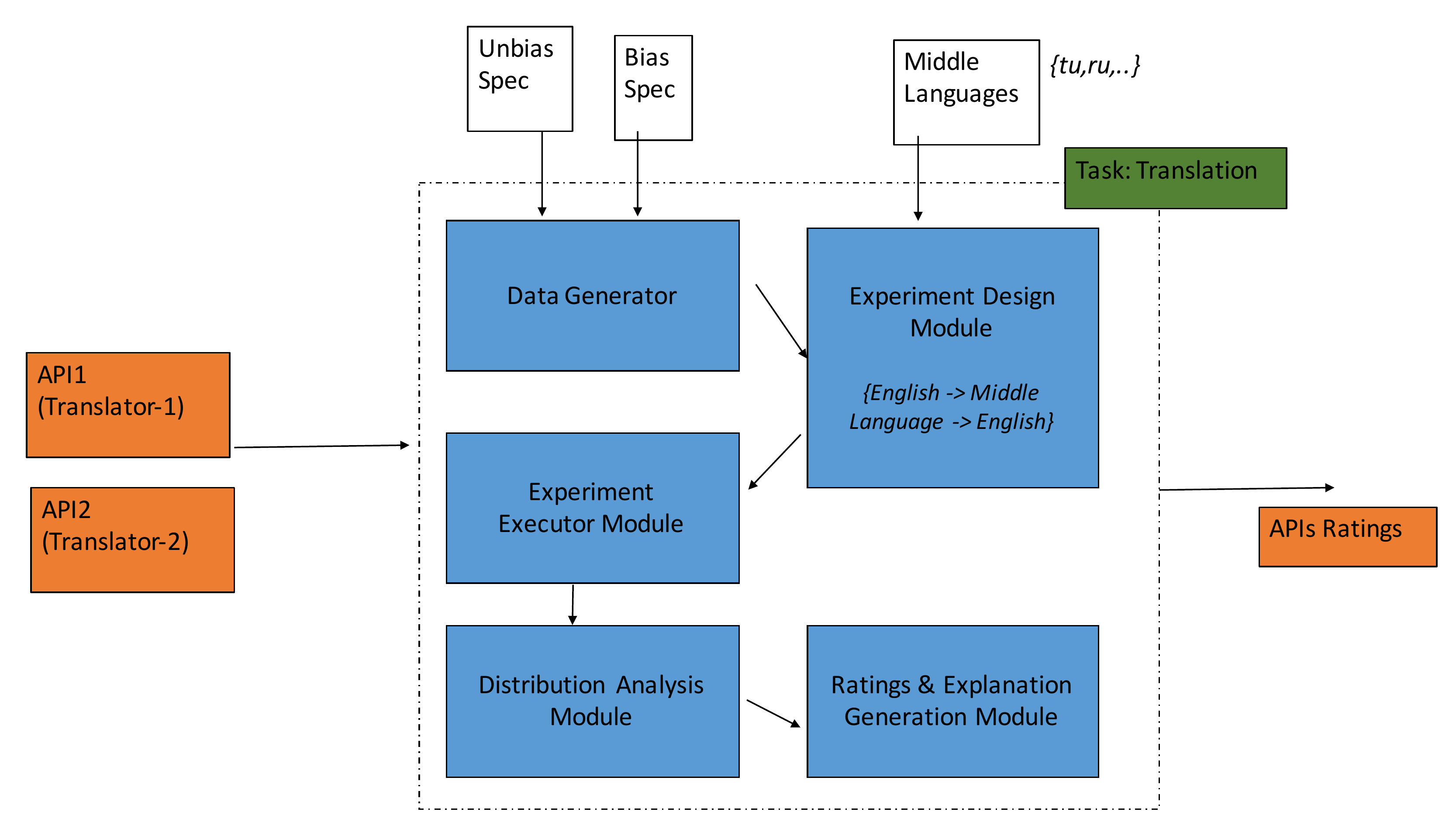}
    \caption{Architecture of Implementation.}
    \label{fig:arch}
\end{figure}

We implement the above described rating procedure 
to rate translation services by using the service twice,
from English to a middle language, and from the middle language 
back to English. 
The architecture of the system we implemented is shown in Figure~\ref{fig:arch}. Its input consists of
\begin{itemize}
\item the specification of the API to be rated
\item the biased and unbiased distributions
\item the middle language(s) to be used
\end{itemize}
while the output is the bias rating for the API, that is, 
BS, DSBS, or UCS.

The internal architecture consists of the following modules:
\begin{itemize}
\item The Data Generator module. This module generates data based on specified attribute/value distribution specifications (for biased and unbiased data),
\item The Experiment Design module. In this paper, for the translation task, the above described USRTest is the experiment under consideration.
\item The Experiment Executor module. This module executes the API and collects  testing results.
\item The Distribution Analysis module. This module compares distributions for biased or unbiased content. In particular, the API's output is compared to the specified unbiased and biased data distributions. In this module, we can select suitable tests based on data distributions, e.g., Chi-squared, Kolmogorov Smirnov (K-S) and Kullback–Leibler (K-L) divergence. For our implementation and experiments, we used Chi-squared test as implemented in Apache's Common Maths library\cite{cm}.
\item The Rating and Explanation Generation module. This is the module that produces the rating of the API. In the USRTest case, the module generates ratings considering API's output for different middle languages, data specifications, and an aggregation criterion to combine the rating results of the various test into an overall rating.
\end{itemize}


\section{Experimental Results}

\subsection{Choice of AI Translation Services and Middle Languages}

As mentioned earlier, we consider two text translation services (AI services):  Google Translate \cite{google-translate,google-translate-tr} ($A^t_g$) and Yandex Translate \cite{yandex-translate,yandex-translate-tr}($A^t_y$). Using these translation services, we translate a text from English to a middle language, and then from the middle language back to English. In selecting translators, we had to take care of a number of practical issues: supported API security mechanisms and whether they can allow repeated calls to the service in a short period for experiments, invocation cost, supported languages and availability of documentation about their usage and statistical working\cite{google-translate-tr,yandex-translate-tr}. 

The 8 middle languages ($M_i$) we consider are  supported by both translators: Arabic (ar), French (fr), Hindi (hi), Italian (it), Portuguese (pt), Russian (ru), Spanish (es) and Turkish (tr). The experiments can be extended seamlessly to other translators and their supported middle languages.

\subsection{Data Generation}

The text to be translated is made up of two sentences containing one gender place-holder each. Its format is: \\
{\em $\prec$Gender$\succ$ is a $\prec$Occupation-Performer$\succ$. $\prec$Gender$\succ$ is a $\prec$Occupation-Performer$\succ$.} 
We chose this two-sentence format because we wanted a text that could include both genders. This allows us to expose in a more articulate way the possible bias translation issues\footnote{In general, the data generator will create sentences of length equal to the number of non-trivial values an attribute of interest can take.}.

For the gender, we use either He or She. For the occupation, we use a list of occupations from a public site\footnote{http://www.vocabulary.cl/Basic/Professions.htm}. An example is - {\em She is a Florist. He is a Gardener.}

We call a block of input that has to be translated from English $\rightarrow M_i \rightarrow$ English, where $M_i$ is the middle language, as a data block. It consists of 20 texts in above format in our experiments.  

Given a selection of unbiased and biased gender distributions, we generate unbiased and biased data blocks. 
Since we have three choices for gender (He, She, and Other), 
a distribution is a triple (x,y,z), where $x+y+z = 1$ and where $x$ is the percentage of occurrences of He, $y$ is the percentage of occurrences of She,
and $z$ is the percentage of occurrences of Other.
For unbiased data, we use the distribution (0.5, 0.5, 0.0): 50\%
of occurrences should be He, 50\% should be She, and none should be Other. 
For biased data, we use the distributions (0.1, 0.9, 0.0) and (0.9, 0.1, 0.0). The fact that {\em Other} is nil (0.0) in data specifications means that we never generate it as part of the input data (that is, only He or She appear in the input data). However, Other may appear in the output data because of the text that a translation service generates. Although we report experiments with a particular setting,  the system can work with any number of biased and unbiased distribution declarations, and we indeed experimented with  other choices.

Figure~\ref{fig:example} showed an example of generated input text and USRTest responses for both translators and all middle languages.

\subsection{Experimental Setting}

For each translator and middle language, we generate 3 data blocks: 1 unbiased and 2 biased, following the distributions defined above. So, we generate 3 * 20 = 60 texts and 2*60 = 120 translations (one from English to $M_i$, and one from $M_i$ to English). We then run the experiment for 8 $M_i$ and 2 translators for a total of 120 * 8 * 2 = 1920 translations.

\begin{figure}[ht]
    \centering
   \includegraphics[width=0.45\textwidth]{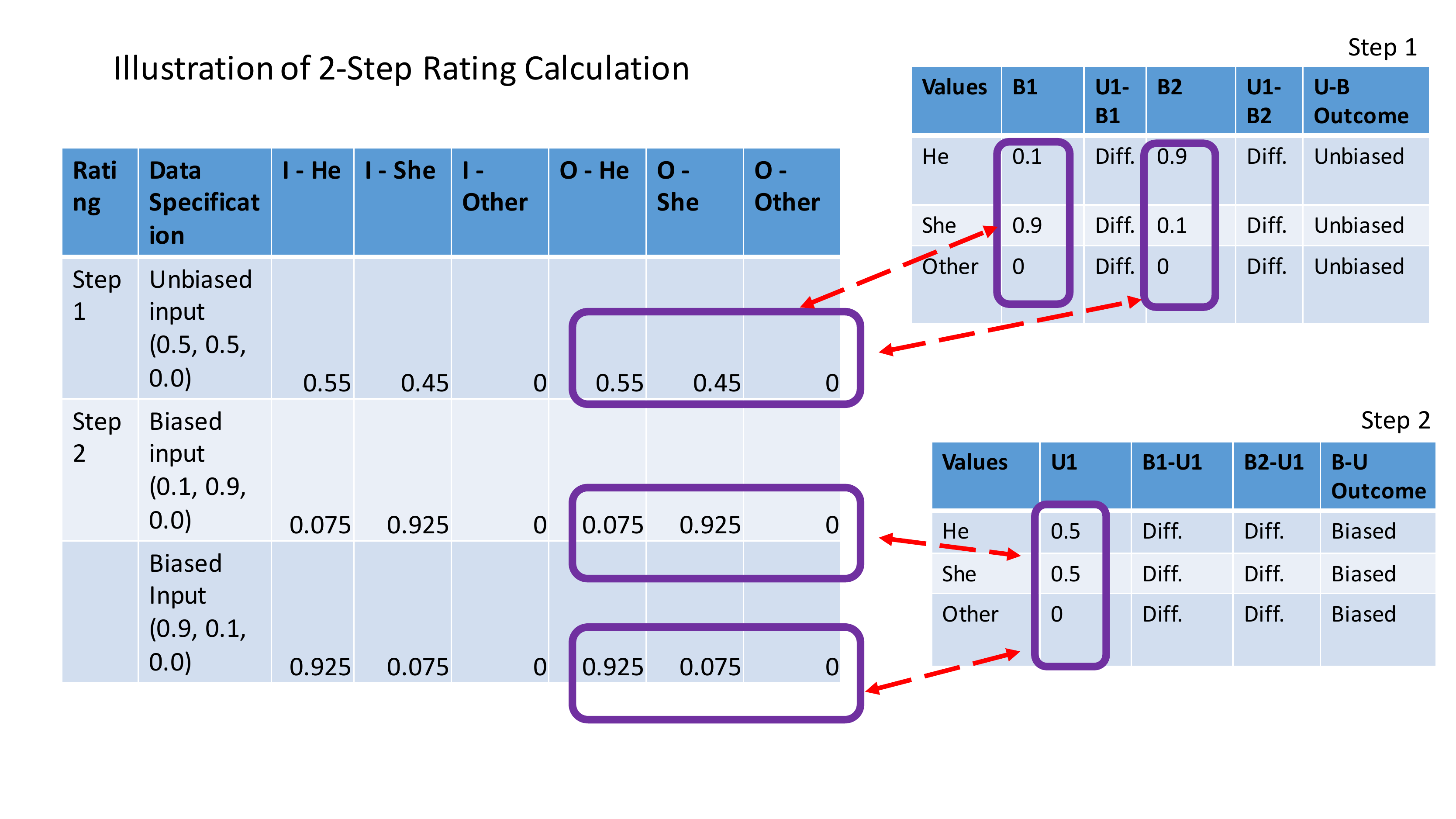}
    \caption{Illustration of rating calculation for $M_i$ = Spanish for $a^t_g$. I-* refer to inputs and O-* refer to outputs. The shown output distribution is average over the data block. The rating generated is DSBS.}
    \label{fig:calc}
\end{figure}

In Figure~\ref{fig:calc}, an illustration of rating calculation is shown for Google translator service($A^t_g$) with Spanish (es) as the middle language ($M_i$). Each row in the left table shows the service's average performance for a data block for unbiased (row 2) and biased (rows 3 and 4) input distributions. The output is then compared to biased distribution specifications (upper right table) in Step-1 and unbiased specifications (lower right table) in Step-2 using Chi-squared two sample test at 95\% confidence\footnote{Note: Chi-squared works with actual counts of results and not distributions. We do the necessary conversions using data block size.}. The criterion to aggregate results is worst case (i.e., boolean or)\footnote{Voting is an attractive alternative  to explore if there are many specifications.}. Thus, if at least one of the distribution comparisons returns similarity, the full comparison is considered to return similarity. 
The similarity is then used to judge whether the output distribution is declared to be biased or not, by comparing it to the declared biased or unbiased distributions.





\subsection{Results and Discussion}

Table~\ref{tab:summary} shows the results of our experimental analysis. Each row shows the performance on a translator on USRTest task for the given middle language. The last two rows show an aggregate rating considering worst-case performance (UCS  $\succ$ DSBS $\succ$ BS, where $\succ$ denotes preferred fairness behavior).

We notice that $A^t_g$ is biased for Turkish and Italian, UCS for Hindi and DSBS for the other middle languages. 
On the other hand, $A^t_y$ is biased for Hindi and DSBS for the other middle languages. 

To better analyze the results, we should consider the nature of the middle languages. We note that both Turkish and Hindi have a single word referring to gender in 3rd person. So, when English text is translated to these languages, the distinction of gender is lost, and while translating back to English, of course the translators have problem recovering them. 
However, other middle languages have separate words for 3rd person gender reference. This is the case of Italian, for which $A^t_g$ seems not able to distinguish gender. However, in Italian, sentences without a subject (thus not explicitly stating if gender is He, She, or something else - Other) are grammatically valid. So, once such a sentence is generated, it is not surprising that the translation service cannot recover the gender.
However, the translation should not lose the gender when translating from English to Italian, which instead happens for some of the data blocks.

While the results and ratings are interesting and insightful, they should only be considered preliminary and a prototypical validation of the rating approach. A definitive rating of AI (translator) service needs to consider all supported middle languages, different forms of input text, different data block sizes and multiple measures of comparing distributions.
However, despite the limitations,  services can be tested  by a 3rd party, rated along the presented bias scale, and advertised on API service catalogs as part of meta-data so that API consumers are aware of  potential bias implications of services they re-use.

 
\begin{table}
\centering

    \begin{tabular}{|l|l|l|l|}
    \hline
          No. & Middle Lang. & Rating (G) & Rating (Y)  \\ \hline \hline
     1.    & Turkish (tu) & BS & DSBS  \\ \hline
     2.    & Russian (ru) & DSBS & DSBS  \\ \hline
     3.    & Italian (it) & BS &  DSBS \\ \hline
     4.    & Spanish (es) & DSBS &  DSBS \\ \hline     
     5.    & Hindi (hi) & UCS &  BS \\ \hline
     6.    & Portuguese (pt) & DSBS &  DSBS \\ \hline
     7.    & French (fr) & DSBS &  DSBS \\ \hline
     8.    & Arabic (ar) & DSBS & DSBS  \\ \hline \hline
           & Overall &  BS & BS \\ \hline
           & Overall- (excluding  &  BS & DSBS \\ 
           & Tu, Hi) &   &  \\ \hline    
    \end{tabular}
    
    \caption{Ratings of translation APIs. }
    \label{tab:summary}
\end{table}



\section{Rating Composite AI Services}

To build large, complex AI applications, APIs are commonly  composed  sequentially.
As an example, consider a service that takes an image and generates a sentiment, which can be obtained by composing sequentially a service that takes an image and generates a caption, and another service that takes the caption and generates a sentiment.
If we know how to rate each of the two services against bias, 
we would like to be able to compose these ratings to generate the ratings of the composite service, without having to start from scratch.

We consider the task of rating composite AI services of the form $A_i \star A_j$, where $\star$ is sequential composition.  As shown in Table~\ref{tab:seq-comp}, if a biased system $A_i$ is sequentially composed with another biased system $A_j$, the outcome of the composite service can be biased (BS), unbiased compensation (UCS) or data-sensitive biased (DSBS). In fact, the two biased system could compensate each other, or they could just follow the same biased pattern. So this is the most unfortunate case in which the composite system has to be tested anew as a separate entity. 

If instead the second service is biased (BS), but the first one is not, then the composite service is biased (BS), since it follows the bias behavior of the biased service. The same happens when the first service is biased (BS) and the second one follows the data (DSBS).
If instead the first service is biased (BS) but the second one can compensate bias (UCS), the overall service cannot see any bias (UCS).
Finally, when the first service is unbiased (UCS), or follows the bias of the data (DSBS), the composite system follows the behavior of the second service.


To summarize:
\begin{enumerate}
\item A BS $\star$ BS service can behave in all possible ways related to bias.
\item  A [.] $\star$ UCS service will compensate for bias in the first service.
\item A DSBS $\star$ [.] service reflects the characteristic of the second service.
\end{enumerate}

Thus, apart from one case, we can avoid rating the composite service and rather exploit the existing rating of each of the two components. 
In fact, if we start by rating the second service, and it turns out to be able to compensate bias (UCS), we can immediately infer that the overall system is UCS too.


 
\begin{table}
\centering
    \begin{tabular}{|l|l|l|l|}
    \hline
          $A_i$ $\star$ $A_j$ & BS & UCS & DSBS  \\ \hline \hline
     BS   & BS/ UCS/ DSBS & UCS & BS  \\ \hline
     UCS    & BS & UCS & DSBS  \\ \hline
     DSBS    & BS & UCS &  DSBS \\ \hline                               
    \end{tabular}
    \caption{Sequential Composition of APIs. The labels of the rows are for the first service, while the labels of the columns are for the second service.}
    \label{tab:seq-comp}
\end{table}


\section{Conclusions and Future Work}

As AI is increasingly getting pervasive in our personal and professional life, concerns are raised on the ethical behavior of  services that
are based on AI technologies. In this paper, we considered the ability of an AI system to behave fairly, that is, not showing bias against any part of the values (population) of entities affected by the AI decisions. In particular, we looked at rating the possible bias exposed in the behavior of translator AI services for which the consumer does not have access to the service's training dataset. 

We envisage a 3rd party rating agency that is independent of the AI API producer or consumer and has its own set of biased and unbiased data, with customizable distributions.
We proposed a 2-step rating approach that generates bias ratings along a 3-level scale signifying whether the AI service is unbiased compensating, data-sensitive biased, or biased.

While our approach is general and does not depend on the kind of data or algorithm used by the AI service, we focused on text dataset and translation services to perform an extensive experimental analysis. Our experiments analyzed two translation services (Google translate and Yandex) and used eight middle languages (since we translate from English to a middle language and then back to English, to compare input and output sentences) to reveal interesting, but preliminary insights. 
We also discussed the possible modular composition of 
the bias behavior of simple service that are composed sequentially to generate more complex services, showing that in most cases, it is possible to exploit the bias rating of the components to rate the composite service.

We believe that our procedure can be very useful to check and assess
the bias behavior of AI services. This is needed since
AI services are increasingly used and there is currently no existing way to know whether they are biased or not. 


We envision several lines for future work.
First, our procedure considers distributions over nouns (He, She, and Other) but not over other linguistic characteristics like verbs. Some languages have different verb forms based on 
gender and they can be considered to estimate gender distortion.
Second, the procedure is run on 8 middle languages but it
      can be extended to all languages supported by
      the considered translators. Similarly, only two translators were 
      considered but the experimental work can be extended to others.
      Also, a more detailed analysis of the role of the middle languages can help assess the bias behavior of the translation service.
The experiments were limited to gender test; other attributes like race, places of workshop, can be considered.
Finally, the experiments were conducted with one form of generated text
(two sentences with similar structure). Alternatively, more complicated sentences can be generated for testing, in order to be more aligned with the kind of text that people actually use in language translators.




\bibliographystyle{ACM-Reference-Format}
\bibliography{ethics}

\end{document}